\documentclass[sigconf]{acmart}
\usepackage[utf8]{inputenc}
\usepackage{algorithmic}

\usepackage[linesnumbered,lined,boxed]{algorithm2e}
\usepackage{verbatim}
\usepackage{xcolor}
\usepackage{float}
\usepackage{tabularx} 
\usepackage{booktabs}
\usepackage{multirow}
\usepackage{mathtools}
\usepackage{setspace}
\usepackage{placeins}
\usepackage{array}
\usepackage{booktabs}
\usepackage{makecell, booktabs, caption}
\usepackage{footnote}   
\usepackage{tikz}
\usepackage{pgfplots}
\usepackage{caption}
\usepackage{subfig}

\usepackage{amsmath}
\usepackage{algorithmic}

\usepackage{mathtools}
\usepackage{setspace}
\usepackage{placeins}
\usepackage{caption}
\usepackage{subfig}
\usepackage{pgfplots}
\usetikzlibrary{matrix}
\usepgfplotslibrary{groupplots}

\usepackage{xcolor,pifont}

\newtheorem{definition}{Definition}

\usepgfplotslibrary{statistics}

\usepackage{xcolor,pifont}

\definecolor{cycle2}{RGB}{106, 191, 0}
\definecolor{cycle3}{RGB}{191, 0, 0}

\definecolor{amber}{rgb}{1.0, 0.75, 0.0}
\definecolor{awesome}{rgb}{1.0, 0.13, 0.32}
\definecolor{ao(english)}{rgb}{0.0, 0.5, 0.0}

\title{MOOMIN: Deep Molecular Omics Network for Anti-Cancer Drug Combination Therapy}
\author{Benedek Rozemberczki}
\affiliation{
  \institution{AstraZeneca}
  \country{Cambridge, United Kingdom}
}

\author{Anna Gogleva}
\affiliation{
  \institution{AstraZeneca}
  \country{Cambridge, United Kingdom}
}

\author{Sebastian Nilsson}
\affiliation{
  \institution{AstraZeneca}
  \country{Gothenburg, Sweden}
}

\author{Gavin Edwards}
\affiliation{
  \institution{AstraZeneca}
  \country{Cambridge, United Kingdom}
}

\author{Andriy Nikolov}
\affiliation{
  \institution{AstraZeneca}
  \country{Cambridge, United Kingdom}
}

\author{Eliseo Papa}

\affiliation{
  \institution{AstraZeneca}
  \country{Cambridge, United Kingdom}
}

\begin{document}

\begin{abstract}
We propose the molecular omics network (MOOMIN) a multimodal graph neural network used by AstraZeneca oncologists to predict the synergy of drug combinations for cancer treatment. Our model learns drug representations at multiple scales based on a drug-protein interaction network and metadata. Structural properties of compounds and proteins are encoded to create vertex features for a message-passing scheme that operates on the bipartite interaction graph. Propagated messages form multi-resolution drug representations which we utilized to create drug pair descriptors. By conditioning the drug combination representations on the cancer cell type we define a synergy scoring function that can inductively score unseen pairs of drugs. Experimental results on the synergy scoring task demonstrate that MOOMIN outperforms state-of-the-art graph fingerprinting, proximity preserving node embedding, and existing deep learning approaches. Further results establish that the predictive performance of our model is robust to hyperparameter changes. We demonstrate that the model makes high-quality predictions over a wide range of cancer cell line tissues, out-of-sample predictions can be validated with external synergy databases, and that the proposed model is data efficient at learning.
\end{abstract}
\maketitle

\section{Introduction}
A large number of human diseases are treated more effectively by the concurrent use of multiple drugs together in \textit{combination therapy} \cite{chesney1999adherence,mokhtari2017combination}. It is a particularly powerful approach in combating cancer, bacterial and viral infections \cite{khdair2010nanoparticle, pennings2013hiv, fair2014antibiotics} which can develop resistance when treated by \textit{monotherapy} -- using a single drug. The application of monotherapy in oncology has several downsides in addition to the development of resistance: high toxic dosages of the drug \cite{demanes2011high}, long-term side effects and poorly targeted treatment which destroys healthy cells \cite{partridge2001side}. In contrast, combination therapy requires lower dosages \cite{albain2008gemcitabine}, the drugs can have synergies at combating cancer \cite{blagosklonny2004analysis} and the intelligent targeting of multiple biological pathways can lead to increased efficacy and reduced toxicity \cite{yap2013development}. Hence, finding synergistic drug combinations in oncology is a highly relevant healthcare problem.

\begin{figure}[h!]
\centering 
\includegraphics[scale=0.15]{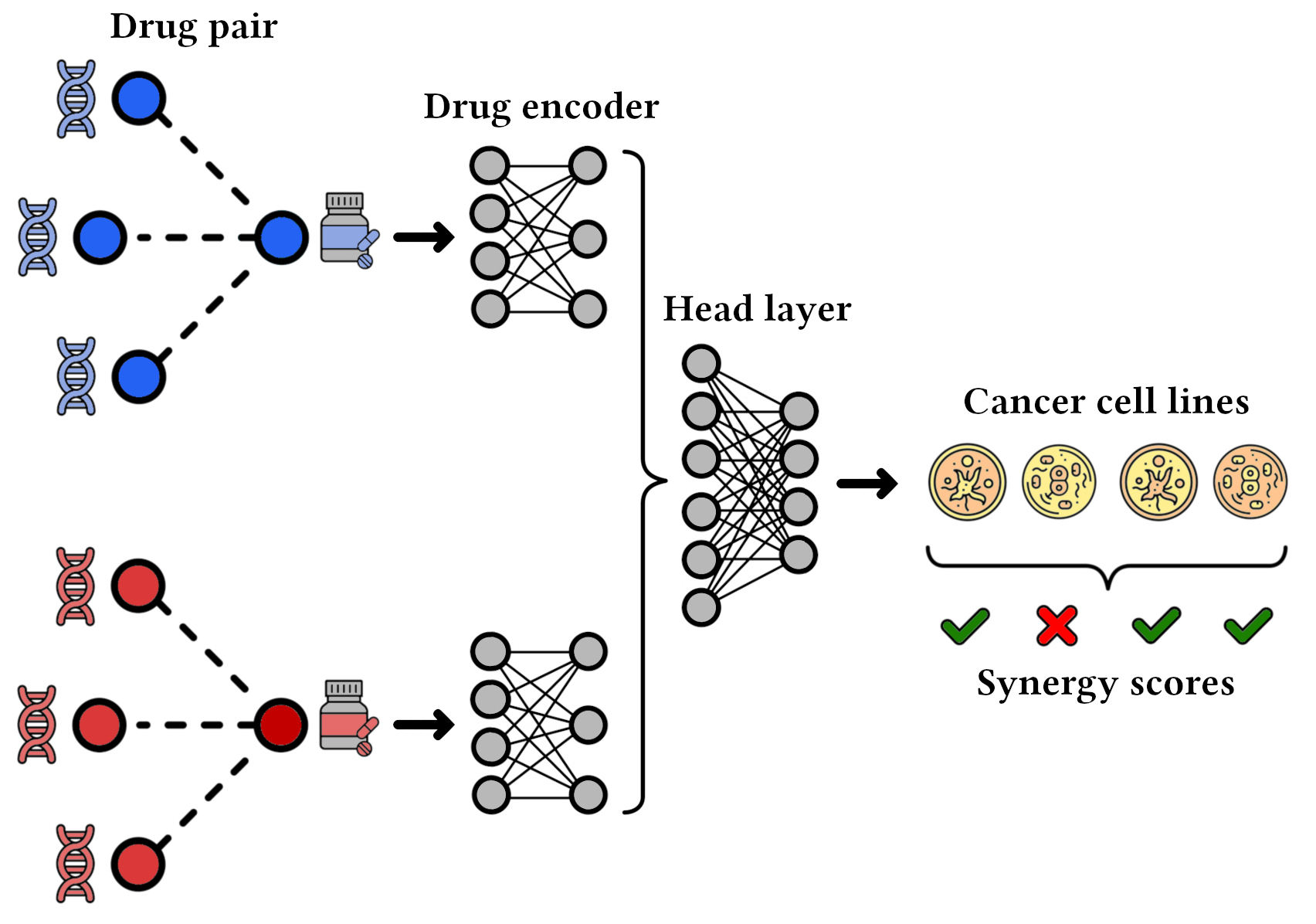}
\caption{Given two drugs (red and blue nodes) MOOMIN represents drugs using protein and compound structures that appear in the neighbourhood of the pair in an interaction graph. Using the representations its head predicts synergistic pairs which are effective at combating cancer.}
\end{figure}

The manual design of effective combination therapy regimes for cancer therapy is a non-trivial task. Even on a single targeted cancer cell type, the number of potential drug pairings (the simplest type of combinations) is quadratic in the number of compounds. This means that only a limited number of potential candidate pairs can be tested by in-vitro and in-vivo experiments. In addition, the experimental evaluation of combinations is hindered by drug and cell line availability \cite{ianevski2019prediction}, conflicting experimental results \cite{liu2020drugcombdb} and the fact that certain outcomes (e.g. polypharmacy side effect) are not the primary interest of experiments \cite{zitnik2018modeling}. Moreover, the synergistic effect of drug pairs depends on the context in which they are applied -- the biological pathways the drugs interact with and the cancer cells being targeted \cite{lehar2009synergistic}. Finally, drug combination experiments are done in a sparse and skewed manner: a limited number of drug pairs are tested on a few cells out of the combinatorial range of possibilities \cite{lehar2009synergistic,liu2020drugcombdb}.  A machine learning-based approach that pre-screens drug combinations would be valuable, shortening the time required to find therapeutic combinations to use in the clinic.

\textbf{Present work.} In this paper we introduce the molecular omics network (MOOMIN), a semi-supervised graph neural network for predicting the synergistic nature of drug pairings for combating cancer cells. We design a custom-tailored multi-resolution multimodal feature extraction model which contextualizes drugs based on molecular features and structural properties of proteins that they interact with in a bipartite graph. Contextualization happens by traversing a drug-protein interaction graph at multiple scales with truncated random walks \cite{grover2016node2vec}. Using the extracted multi-resolution representations we define a scoring layer that can output synergy scores for pairs of drugs and a targeted cancer cell line. The components of the model are trained end-to-end by maximizing the likelihood of correct synergy predictions.

Our main contribution is the definition of multi-scale multimodal drug representations. These learned features integrate molecular information about a given drug and about the drugs and proteins that are in the neighborhood of the source drug at multiple proximity scales in a drug-protein interaction graph. By using a bipartite interaction graph for modeling, the information about drugs and proteins at multiple scales is distilled by separate blocks of features. Therefore, the synergy scoring performance lift gained by the inclusion of multimodal drug-protein information is quantifiable with our model. We also propose an algorithm to approximately contextualize the drugs. This allows scalable model training and efficient synergy scoring inference when a dimension of the dataset such as the number of drugs considered is large. 

The empirical evaluation of our model focuses on the synergy scoring task: given a pair of drugs and a cancer cell line the models have to predict whether the drug combination is synergistic. We compare the predictive performance of MOOMIN with state-of-the-art graph fingerprinting, node embedding, tensor factorization, and deep learning techniques. Using a real-world drug pair synergy database \cite{liu2020drugcombdb} we show that MOOMIN outperforms competing methods by as much as 7.2\%  and 2.5\% in terms of F$_1$ and PR AUC scores on the test set. The predictive performance of MOOMIN is robust across cancer cells conditioned on source tissues and competitive with other methods with a fraction of training data. Out-of-sample predictions made by AstraZeneca scientists on standard drug combination databases demonstrate that MOOMIN can identify novel synergistic drug combinations of approved drugs.

\textbf{Our contributions.} Overall the main contributions of our paper can be summarized as the following:
\begin{itemize}
    \item We propose \text{MOOMIN} an empirically motivated multimodal graph neural network that can predict synergistic drug pairs using protein and molecule features.
    \item We design an efficient mini-batch training algorithm to scale the training and inference of \text{MOOMIN} to large-scale drug synergy prediction problems.
    \item We support evidence using real-world data that \text{MOOMIN} can effectively find synergistic drug pairings. 
\end{itemize}

The remainder of this paper has the following structure. We overview the related work about network representation learning and machine-aided combination therapy in Section \ref{sec:related_work}. We define relevant theoretical concepts, introduce the datasets used in the experiment and show the empirical regularities motivating our model design in Section \ref{sec:preliminaries}. The details of the model are discussed in Section \ref{sec:model}. We experimentally evaluate the model in Section \ref{sec:experiments} on synergy prediction tasks. The paper concludes with Section \ref{sec:conclusions} where we summarize our main findings and point out directions for future research.

\section{Related work}\label{sec:related_work}
Our work touches on machine learning approaches to identify combination therapies as well as various existing graph representation learning techniques at the node and graph levels. We are going to discuss these methods and contrast MOOMIN based on various desired characteristics of the models.
\subsection{Machine learning for combination therapy}
The application of machine learning to combination therapy is commonly done by solving a scoring problem: given a pair of drugs (combination) we want to predict the probability that the pair has a property. The predicted property can be a certain side-effect caused by the combination \cite{zitnik2018modeling, xu2020tip}, an interaction of drugs \cite{kwon2017deepcci,xu2019mr} or the synergistic behavior of the pair \cite{kuenzi2020predicting,preuer2018deepsynergy}. We summarized recent algorithmic developments on these tasks in Table \ref{tab:related_drugs}.

Existing models are differentiated by the task solved and whether predictions can be made about drugs that are not present at training time (induction). Another differentiating factor is the type of input data ingested by the models. Most models exploit heterogeneous graphs with various node types (e.g. drugs, proteins) and structural features of drugs and proteins. On the synergy scoring task MOOMIN is the first one to fuse drug-protein interaction information with actual structural properties of drugs and proteins. 

\begin{table}[h!]
\centering
\footnotesize
\setlength{\tabcolsep}{2pt}
\renewcommand{\arraystretch}{1.2}
\caption{A machine learning task, induction and input data type based comparison of MOOMIN and existing deep learning models that solve combination therapy problems.}\label{tab:related_drugs}
\begin{tabular}{@{} c c c cc cc }

& && \multicolumn{2}{c}{\textbf{Node Types}} & \multicolumn{2}{c}{\textbf{Node Features}}\\ \hline
\textbf{Method} &\textbf{Task}&\textbf{Inductive}&

\textbf{Drug}&\textbf{\,\,Protein\,}&

\textbf{Drug}&\textbf{\,\,Protein\,\,}\\

\hline
SkipGNN   \cite{huang2020skipgnn}      &Side Effect        &&$\bullet$&  &&\\
DECAGON   \cite{zitnik2018modeling}    &Side Effect         &&$\bullet$&$\bullet$&&\\
ESP       \cite{burkhardt2019esp}      &Side Effect        &&$\bullet$&$\bullet$ &&\\
TIP       \cite{xu2020tip}             &Side Effect        &&$\bullet$&$\bullet$ &&\\
DeepDDI   \cite{ryu2018deep}       &Side Effect       &&$\bullet$&              &$\bullet$&\\
SumGNN \cite{yu2021sumgnn}       &Side Effect   &&$\bullet$&$\bullet$ &$\bullet$&\\
\hline
DeepCCI   \cite{kwon2017deepcci}       &Interaction        &$\bullet$&&&           $\bullet$&\\
MR-GNN   \cite{xu2019mr}       &Interaction        &$\bullet$&&&                   $\bullet$&\\
\hline
DrugCell  \cite{kuenzi2020predicting}  &Synergy &$\bullet$&&$\bullet$  &$\bullet$&\\
DeepSynergy   \cite{preuer2018deepsynergy}       &Synergy        &$\bullet$&&     &$\bullet$&\\
MOOMIN (ours)   &Synergy        &$\bullet$&$\bullet$&$\bullet$&           $\bullet$&$\bullet$\\
\bottomrule
\end{tabular}

\end{table}
\vspace{-5mm}

\subsection{Node and graph representation learning}
The architecture proposed in our work distills representations from graphs both at the node and graph level (interactome and molecules). To position our work we give a general overview of machine learning techniques that can learn these representations.

\subsubsection{Proximity preserving node representation learning} Node embedding techniques in this general category map nodes of a graph into a Euclidean embedding space where \textit{pairwise proximity} between nodes is approximately preserved  \cite{tsitsulin2018verse}. Models are differentiated from each other by the notion of proximity such as truncated random walk transition probabilities \cite{perozzi2014deepwalk,grover2016node2vec, rozemberczki2018fast, qiu2018network}, neighbourhood overlap \cite{ahmed2013distributed, tang2015line} or personalized PageRank \cite{ou2016asymmetric}. Importantly encoding nodes in this embedding space does not optimize for the preservation of structural properties directly \cite{henderson2011s, henderson2012rolx,ribeiro2017struc2vec,rossi2020proximity}, but this information can be correlated with the location in the embedding space. Using the node embeddings as input features for a downstream model allows various supervised machine learning tasks such as node classification.

\subsubsection{Graph fingerprints} These statistical fingerprinting techniques create embeddings of whole graphs which preserve \textit{pairwise structural similarities}. Graph fingerprints are commonly created by the factorization of graph-structural feature matrices \cite{narayanangraph2vec,chen2019gl2vec}, scattering transforms \cite{gao2019geometric, verma2017hunt}, spectral feature extraction \cite{tsitsulin2018netlsd, tsitsulin2020just, galland2019invariant} and structural property summarization \cite{chai2018,rozemberczki2020characteristic}. The fingerprints created are fed to a downstream model which ingests these features to predict graph level properties. 
\subsubsection{Inductive node representation learning} In contrast with proximity-based node embedding techniques these methods map nodes into an embedding space based on the similar distribution of vertex attributes in neighborhoods \cite{hamilton2017representation,hamilton2020graph}. Inductive node representation learning techniques can be seen as methods that perform \textit{neural message passing} \cite{gilmer2017neural, battaglia2018relational}. In this paradigm nodes in the graph generate hidden representations with a parametric trainable function and the messages are aggregated in neighborhoods using a weighting scheme and an update function. Models are differentiated from each other based on the message generating function and the message aggregation scheme applied. Messages can be generated by linear models \cite{wu2019simplifying}, single hidden layer  \cite{kipf2017semi, defferrard2016convolutional} or deep neural networks \cite{predict2019klicpera,bojchevski2020scaling}. Aggregation weights can be defined based on exact \cite{kipf2017semi, wu2019simplifying, chiang2019cluster} or sampled  adjacency relations \cite{hamilton2017inductive,ying2018graph}, personalized PageRank \cite{predict2019klicpera,bojchevski2020scaling} or neural network parametrization \cite{graph2018velickovic,rozemberczki2021pathfinder}.

\subsubsection{Pooling for graph representations} Node level representations outputted by graph neural networks can be pooled by permutation invariant functions to generate whole graph representations. This permutation invariant function can be a neural network itself which can perform node sorting \cite{zhang2018end}, top-k node retrieval \cite{gao2019graph,cangea2018towards,knyazev2019understanding} or calculate a weighted average of representations using an attention mechanism \cite{knyazev2019understanding,lee2019self,ying2018hierarchical} to generate the graph level representations.

\section{Preliminaries}\label{sec:preliminaries}
Our work considers contextualizing drugs at multiple scales by structural and molecular properties of compounds and proteins to predict synergetic pairings of drugs. The introduction of a machine learning model which can achieve this requires an appropriate notation of related concepts and a discussion of the datasets integrated.

\subsection{Formal definitions}
The architecture design description extensively uses the definition of drug  set $\mathcal{D}=\{d_1,\dots,d_n \}$, cancer cell line set $\mathcal{C}=\{c_1,\dots, c_k\}$ and protein set $\mathcal{P}=\{p_1,\dots,p_m \}$. Introducing the sets of drugs that can be used to combat cancer cells, proteins that interact with these drugs, and cancer cell lines allows us to formulate the definition of a drug pair synergy set which is our primary interest.

\begin{definition} \textbf{Drug pair synergy set.} A drug pair synergy set defined on $\mathcal{D}$ and $\mathcal{C}$ is the set $\mathcal{S}$ containing tuples of the form $(d,d',c,y)$ where $d,d'\in \mathcal{D}$, $c\in \mathcal{C}$ and $y\in \{0,1\}$ is the synergy label of the drug pairing. 
\end{definition}
The entries in the drug pair synergy set are pairs of drugs which are known to be synergistic or antagonistic at combating certain cancer cells. A positive label means that the pair of drugs are more effective together at fighting the cancer cell, while a zero label shows a lack of synergy.

\begin{definition} \textbf{Drug pair synergy predictor.}
A drug pair synergy predictor defined on $\mathcal{D}$ and $\mathcal{C}$ is the function $\hat{y} = h(d,d',c)$ where $d,d'\in \mathcal{D}$, $c\in \mathcal{C}$ and $0\leq \hat{y}\leq 1$ is the predicted synergy score of the drug pair. 
\end{definition}
A drug pair synergy predictor is a function that predicts the probability of the pair being synergistic at combating a cancer cell. 

\begin{definition} \textbf{Drug-protein interaction graph.} The interaction graph defined on $\mathcal{D}$ and $\mathcal{P}$ is the bipartite graph $\mathcal{G}=(\mathcal{V},\mathcal{E})=(\mathcal{D}\cup \mathcal{P},\mathcal{E})$ where $\mathcal{V}$ and $\mathcal{E}$ are the vertex and edge sets respectively.
\end{definition} 
Our definition of the drug-protein interaction graph does not allow the existence of edges within the drug and protein sets. This hedges against potential data leakage on drug-drug interactions.

\subsection{Dataset}
The dataset \cite{bikg2021} which we used to evaluate MOOMIN was created by the fusion of multiple data sources which we discuss in detail.
\subsubsection{Synergy database} The database which we used comes from DrugCombDB which is a public dataset \cite{liu2020drugcombdb}. After the data fusion, our drug pair synergy set contained 10,911 pairings, 987 unique drugs, and 110 cell lines coming from 14 source tissues. Drug pairs have a positive label when the pair slowed the growth of cancer cells more than what would be expected, based on the effect of the compounds when used alone (synergy), while a zero label is associated with increased cell growth when combined (antagonism).
\subsubsection{Drug-protein interaction graph} The drug and protein interaction network was composed of several sources and has 1,160 drugs and 17,582 human proteins with 232,524 edges between vertices. Drug-protein relations were mainly taken from three public sources: (i) ChEMBL~\cite{gaulton2016chembl}, which included both the verified targets against which the compound was developed as well as interactions discovered experimentally, (ii) CTD~\cite{davis2020ctd}, which aggregates information from research publications, and (iii) Hetionet~\cite{himmelstein2017hetionet}, which itself accumulates a small set of sources to cover all the most relevant domain concepts. These public reference datasets were extended with internal as well as proprietary data from two drug development database products: MetaBase \cite{bolser2012metabase} and Pharmaprojects \cite{tao2011pharmaprojects}.

\subsubsection{Molecular descriptors}
The drug molecules in DrugCombDB are represented as SMILES strings \cite{weininger1988smiles}, a common notation for describing molecular graphs in string form. From the SMILES strings, we extracted the molecular structure of the drugs (connectivity graph and atoms). The structure was extracted using PaDELPy \cite{padelpy}, a Python wrapper of the PaDEL-Descriptor \cite{yap2011padel} package. 
\subsubsection{Protein features}

Protein sequence information can be represented by various classes of numeric descriptors. In this study, we used amino acid composition descriptors (dipeptide composition), along with pseudo-amino acid composition descriptors (amphiphilic pseudo-amino acid composition). Amino acid composition descriptors reflect the frequency of dipeptide combinations of amino acids in a protein sequence. Pseudo-amino acid composition descriptors in addition incorporate local properties of protein sequences such as the correlation between residues or a certain distance between groups of residues \cite{chou2001prediction}. All descriptors were computed on a full set of human proteins using the protr package \cite{protr}. 
\subsection{Exploratory observations and motivation}
The effects of drug-drug combinations are widely studied experimentally, the number of parameter combinations to test makes such validation very time-consuming and expensive \cite{liu2020drugcombdb}.
Thus, available experimental data on synergistic and antagonistic effects are necessarily sparse. Moreover, it is hard to extrapolate from existing pairs to new pairs of drugs or new types of cells due to several factors.
\begin{figure}[h!]

	\centering
	\begin{tikzpicture}[scale=0.35,transform shape]
	\tikzset{font={\fontsize{17pt}{12}\selectfont}}
	\begin{groupplot}[group style={group size=2 by 1,
		horizontal sep=60pt, vertical sep=60pt,ylabels at=edge left},
	width=0.63\textwidth,
	height=0.35\textwidth,
	ymin=0.00,
	ymax=1.00,
	legend columns=3,
every tick label/.append style={font=\bf},
    y tick label style={
        /pgf/number format/.cd,
            fixed,
            fixed zerofill,
            precision=0,
        /tikz/.cd
    },
 enlarge x limits=true,
	grid=major,
	grid style={dashed, gray!40},
	scaled ticks=false,
	inner axis line style={-stealth}]

 \nextgroupplot[
   xlabel= Most common cell lines,
    ybar=0pt,
      every tick/.style={
        black,
        semithick,
      },
    bar width=22pt,
    enlargelimits=0.17,
    title=\textbf{(A)},
    ymax=0.5,
    ylabel={Ratio of experiments},
    legend style={at={(0.5,-0.15)},
      anchor=north,legend columns=-1},
yticklabels={0, 0.1,0.2,0.3,0.4,0.5},
ytick={0, 0.1,0.2,0.3,0.4,0.5},
    symbolic x coords={KBM-7,DIPG25,DD2,TC-71,3D7},
    xtick={KBM-7,DIPG25,DD2,TC-71,3D7},
    ]

\addplot [fill=red!55,draw=red,error bars/.cd,y dir=both,y explicit]  coordinates {
(KBM-7,0.378)
(DIPG25,0.0791)
(DD2,0.0614)
(TC-71,0.0591)
(3D7,0.0567)};

 \nextgroupplot[
   xlabel=Most common drugs,
    ybar=0pt,
      every tick/.style={
        black,
        semithick,
      },
    bar width=22pt,
    title=\textbf{(B)},
    enlargelimits=0.17,
    legend columns=4,
    ymax=0.06,
    legend image post style={solid},
    legend style={at={(0.5,-0.25)},nodes={scale=1.5, transform shape}, 
      anchor=north,legend columns=-1},
    ylabel={Ratio of experiments},
yticklabels={0.0,0.02,0.04,0.06},
ytick={0.0,00.02,0.04,0.06},
    symbolic x coords={5-FU,MK-4827,MK-5108,MK-2206,BEZ-235},
    xtick={5-FU,MK-4827,MK-5108,MK-2206,BEZ-235},
    	legend style = { column sep = 10pt, legend columns = 1, legend to name = grouplegend, font=\small}  ]

\addplot+[fill=red!55,draw=red,error bars/.cd,y dir=both,y explicit]  coordinates {
(5-FU,0.0541)
(MK-4827,0.0448)
(MK-5108,0.0341)
(MK-2206,0.0295)
(BEZ-235,0.0209)};

	\end{groupplot}

	\end{tikzpicture}
	
	\caption{Drug combination pairs are frequently (A) tested on popular cell lines and (B) include popular drugs.}
	\label{fig:skewness}
\end{figure}
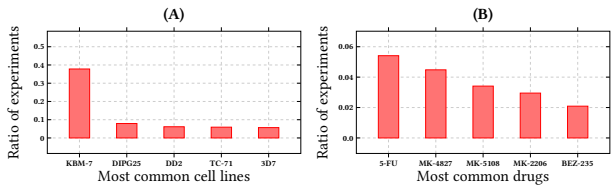
\vspace{-5mm}

\subsubsection{Popular cell lines and drugs.} Experiments are often performed only with popular drugs on popular cell lines, while effects in some tissues are rarely studied. This also makes it difficult to train separate predictive models for different cell lines and drugs. Looking at Figure \ref{fig:skewness} it can be seen that 63\% of experiments in DrugCombDB are done on the 5 most popular cell lines and 18\% of synergy tests are carried out with the top 5 drugs.
\begin{figure}[h!]

	\centering
	\begin{tikzpicture}[scale=0.35,transform shape]
	\tikzset{font={\fontsize{17pt}{12}\selectfont}}
	\begin{groupplot}[group style={group size=2 by 1,
		horizontal sep=60pt, vertical sep=60pt,ylabels at=edge left},
	width=0.63\textwidth,
	height=0.35\textwidth,
	ymin=0.00,
	ymax=1.00,
	legend columns=3,
every tick label/.append style={font=\bf},
    y tick label style={
        /pgf/number format/.cd,
            fixed,
            fixed zerofill,
            precision=0,
        /tikz/.cd
    },
 enlarge x limits=true,
	grid=major,
	grid style={dashed, gray!40},
	scaled ticks=false,
	inner axis line style={-stealth}]

 \nextgroupplot[
   xlabel= Number of tested cell lines,
    ybar=0pt,
      every tick/.style={
        black,
        semithick,
      },
    bar width=18pt,
    enlargelimits=0.17,
    title=\textbf{(A)},
    ylabel={Ratio of conflicting pairs},
    legend style={at={(0.5,-0.15)},
      anchor=north,legend columns=-1},
yticklabels={0, 0.2,0.4,0.6,0.8,1.0},
ytick={0, 0.2,0.4,0.6,0.8,1.0},
    symbolic x coords={2,3,4,5,6,$7\leq$},
    xtick={2,3,4,5,6,$7\leq$},
    ]

\addplot [fill=blue!55,draw=blue,error bars/.cd,y dir=both,y explicit]  coordinates {
(2,0.266)
(3,0.295)
(4,0.575)
(5,0.5)
(6,0.831)
($7\leq$,0.907)};

 \nextgroupplot[
   xlabel=Cell line tissue,
    ybar=0pt,
      every tick/.style={
        black,
        semithick,
      },
    bar width=18pt,
    title=\textbf{(B)},
    enlargelimits=0.17,
    legend columns=4,
    legend image post style={solid},
    legend style={at={(0.5,-0.25)},nodes={scale=1.5, transform shape}, 
      anchor=north,legend columns=-1},
    ylabel={Ratio of conflicting pairs},
yticklabels={0.0,0.2,0.4,0.6,0.8,1.0},
ytick={0.0,0.2,0.4,0.6,0.8,1.0},
    symbolic x coords={Bone,Brain,Breast,Ovary,Prostate,Skin},
    xtick={Bone,Brain,Breast,Ovary,Prostate,Skin},
    	legend style = { column sep = 10pt, legend columns = 1, legend to name = grouplegend, font=\small}  ]

\addplot+[fill=blue!55,draw=blue,error bars/.cd,y dir=both,y explicit]  coordinates {
(Bone,0.032)
(Brain,0.129)
(Breast,0.586)
(Ovary,0.675)
(Prostate,0.352)
(Skin,0.650)};

	\end{groupplot}

	\end{tikzpicture}
	
	\caption{Drug combination pairs often produce opposite effects on different cell lines (synergy vs antagonism). The share of such pairs (A) grows with the number of tested cell lines and (B) varies greatly depending on the tissue.}
	\label{fig:drug_pairs_conflicting}
\end{figure}
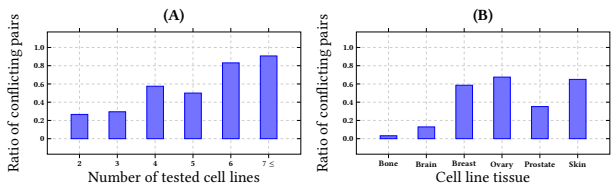
\vspace{-5mm}
\subsubsection{Opposing effects.} In DrugCombDB third of drug pairs that have been tested on different cell lines have shown a synergistic effect on one cell line, but antagonistic on another. As seen in Figure~\ref{fig:drug_pairs_conflicting}, as the same pair of drugs are tested against multiple cell lines, the chance of discovering opposite effects grows: i.e., to predict all effects of applying drugs in combination it is important to take into account the whole range of potentially affected cells and tissues. This problem is made even more complex by the fact that the same pair of drugs often exhibit opposite effects within the same tissue: e.g., this was the case for 67\% of drug pairs tested on ovary cell lines, but only less than 4\% for bone.

\begin{figure*}
\centering 
\includegraphics[scale=0.065]{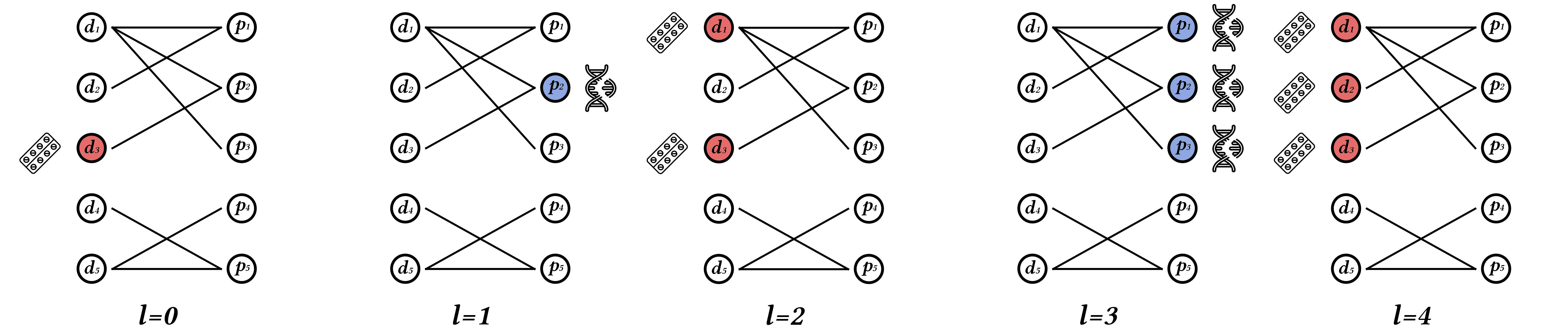}
\caption{In this example the drug $d_3$ has a representation formed by concatenating aggregated drug and protein representations. These representation blocks at even scales $l=0,2,4$ are learned by aggregating the representations output by the drug-encoder using geometric properties of the compounds in the sets $\left\{d_1\right\},\left\{d_1;d_3\right\}$ and $\left\{d_1;d_2;d_3\right\}$. The odd scale representations at $l=1,3$ are learned by combining the protein representations generated by the protein encoder using the structural properties of the proteins in the sets $\left\{p_1\right\}$ and $\left\{p_1;p_2;p_3\right\}$.}\label{fig:main_explanation}
\end{figure*}

\section{The Framework Design}\label{sec:model}
The design of MOOMIN was influenced by the empirical regularities which we discussed in the preliminaries: dataset imbalance, sparsity of the synergy database, and cell line conditional synergistic and antagonistic behavior of drug pairs. 

\subsection{Encoders}
The decision about a pair of drugs being synergistic at effectively shrinking a cancer cell line depends on information about the pair of drugs, protein-drug interactions, and the targeted cell line. 
\subsubsection{Drug encoder} Learning non-contextualized representations of an individual drug $d\in\mathcal{D}$ is handled by the drug encoder $f_{\Theta_D}(\cdot)$ which is parametrized by $\Theta_D$, uses the molecular structure $\mathcal{M}_d$ to generate $\textbf{h}_d$ a vector representation of drugs -- see Equation \eqref{eq:drug_encoder}.
\begin{align}
\textbf{h}_d=f_{\Theta_{D}} (\mathcal{M}_d),\quad \forall d \in \mathcal{D}\label{eq:drug_encoder}
\end{align}
A drug encoder, for example, could use a graph neural network to learn node embeddings of the atoms in the molecule, pool these atom-level representations, and concatenate the pooled representations to create a hidden representation learned from the molecule structure and atom-level features. Our implementation of \text{MOOMIN} used APPNP \cite{predict2019klicpera,bojchevski2020scaling} for learning the atom embeddings. Atom representations were fed to a fully connected feed-forward neural network with two hidden layers and concatenated together after pooling with mean, maximum, and minimum functions to generate molecular descriptors.
\subsubsection{Protein encoder} Creating representations of proteins in the interaction graph is done by the encoder $f_{\Theta_P}(\cdot)$ which is parametrized by $\Theta_P$ and uses structural information $\textbf{x}_p$ to generate $\textbf{h}_p$ a vector representation of the protein $p\in \mathcal{P}$ -- see Equation \eqref{eq:protein_encoder}.
\begin{align}
\textbf{h}_p&=f_{\Theta_P} (\textbf{x}_p), \quad \forall p \in \mathcal{P}\label{eq:protein_encoder}
\end{align}
Protein encoders in \text{MOOMIN} are feedforward neural networks with a single hidden layer and take structural feature vectors of proteins as input. 
\subsubsection{Cell line encoder} The cancer cell lines on which the drugs are tested also need representations which are generated by the cell encoder $f_{\Theta_C}(\cdot)$, parametrized by $\Theta_C$ which uses cancer cell features as input in Equation \eqref{eq:cell_encoder}.
\begin{align}
\textbf{h}_c&=f_{\Theta_C}(\textbf{x}_c),\quad \forall c \in \mathcal{C}\label{eq:cell_encoder}
\end{align}

The cell line encoders were conceptualized as trainable parametric vector embeddings of cancer cells in our implementation of \text{MOOMIN}. This way induction concerning new cell lines is not possible. However, by choosing an encoder that uses exogenous cell features this limitation can be lifted.
\subsection{Multimodal drug representation}
The representations of drugs in combinations are learned from multimodal data and the drug-protein graph. Individual representations of the compounds are contextualized by those protein and drug representations which can be reached at various proximity scale $l\in\left\{0,\dots,r\right\}$ by truncated random walks. In practice this design choice allows the model to learn \textit{difference operators} \cite{abu2019mixhop} at various scales. This idea is described with a toy example in Figure \ref{fig:main_explanation}.
\subsubsection{Contextualization by neighbors at jumps.} The multi-scale representation for drug $d\in\mathcal{D}$ is the vector $\textbf{m}_d$ defined by Equation \eqref{eq:multiscale_probs} -- where $\| $ is the column-wise concatenation operator.

\begin{align}
\textbf{m}_d=&\underset{\mathclap{l \in \{0,\dots ,r\}}}{\Bigg\Vert}\sum\limits_{w \in V}P(v_{j+l}=w|v_j=d)\cdot \textbf{h}_w, \quad \forall d \in \mathcal{D}\label{eq:multiscale_probs}
\end{align}

In Equation \eqref{eq:multiscale_probs} the probability $P(v_{j+l}=w|v_j=d)$ describes the probability that a truncated discrete random walk on $\mathcal{G}$ which started at drug $d$ after $l$ steps terminates at node $w$. The vector $\textbf{h}_w$ is a learned representation of $w\in V$outputted by respectively the drug or protein encoders  described by Equations \eqref{eq:drug_encoder} and \eqref{eq:protein_encoder}. Because $\mathcal{G}$ is bipartite, when the scale is even the walker terminates at drugs, and at an odd scale, it terminates at proteins. In simple terms, $\textbf{m}_d$ is formed by concatenating together proximity-weighted drug and protein representations in an alternating fashion. 

The probability scores in Equation \eqref{eq:multiscale_probs} can be expressed algebraically. Let us denote the row normalized adjacency matrix of $\mathcal{G}$ as  $\widehat{\mathbf{A}}$ and we can reformulate the multi-scale representation definition of drugs as Equation \eqref{eq:multiscale_adjacency}.

\begin{align}
\textbf{m}_d&=\underset{\mathclap{l \in \{0,\dots ,r\}}}{\Bigg\Vert}\sum\limits_{w \in V}
\widehat{\textbf{A}}^l_{d,w}\cdot \textbf{h}_{w}, \quad \forall d \in \mathcal{D}\label{eq:multiscale_adjacency}
\end{align}

The exact calculation of the multi-scale multimodal drug representations described by Equation \eqref{eq:multiscale_adjacency} is not always feasible in practical settings. One limiting condition is an indirect result of the \textit{low effective diameter phenomenon} -- as the length of the truncated random walks being considered is increased the normalized adjacency matrix powers lose sparsity \cite{chakrabarti2006graph}. In practice this means $\mathcal{O}(|V|^3)$ time and $\mathcal{O}(|V|^2)$ space requirement for calculating and storing $\widehat{\textbf{A}}^l$ when $l$ is large. Another considerable issue is the space requirement needed for all of the protein and drug representations when the drug-protein graph is large. 
	\begin{algorithm}[h!]
{\small		\DontPrintSemicolon
		\SetAlgoLined
		\KwData{$\mathcal{G}$ -- Bipartite drug -- protein interaction graph.\\
		$\,\,\,\,\,\,\,\,\,\,\,\,\,\,\,d$ -- Source drug of interest.\\
		$\,\,\,\,\,\,\,\,\,\,\,\,\,\,\,s$ -- Sample size.\\
		}
		\KwResult{$\overline{\textbf{m}}_d$ -- Sampled multi-scale representation of drug $\textit{d}.$}
$\widetilde{\textbf{m}}^0_d\leftarrow f_{\Theta_{D}}(\mathcal{M}_d)$\;
$\overline{\textbf{m}}_d\leftarrow\textbf{0}$\;
\For{$i \in \left \{1,\dots ,s\right \}$}{
$\widetilde{d}\leftarrow d$\;
$\widetilde{\textbf{m}}_d\leftarrow\widetilde{\textbf{m}}^0_d$\;
    \For{$l \in \left \{1,\dots ,r\right \}$}{
            \uIf{$l \mod 2 = 1$}{
            $\widetilde{p} \leftarrow \textbf{Random neighbor}(\mathcal{G}, \widetilde{d})$\;
            $\widetilde{\textbf{m}}^l_d \leftarrow f_{\Theta_p} (\textbf{x}_{\widetilde{p}}) $\;
            $\widetilde{\textbf{m}}_d \leftarrow [\widetilde{\textbf{m}}_d   \|  \widetilde{\textbf{m}}^l_d]$\;
            }
            \Else{
            $\widetilde{d} \leftarrow \textbf{Random neighbor}(\mathcal{G}, \widetilde{p})$\;
            $\widetilde{\textbf{m}}^l_{\widetilde{d}} \leftarrow f_{\Theta_D} (\mathcal{M}_d) $\;
            $\widetilde{\textbf{m}}_d \leftarrow [\widetilde{\textbf{m}}_d   \|  \widetilde{\textbf{m}}^l_d  ]$\;
            }
            }
            $\overline{\textbf{m}}_d \leftarrow \overline{\textbf{m}}_d + \widetilde{\textbf{m}}_d$ \;
}
            $\overline{\textbf{m}}_d \leftarrow \overline{\textbf{m}}_d /s$\;
}
		\caption{Approximate truncated random walk sampled multi-scale drug representation forward pass with MOOMIN.}\label{alg:sampling_algorithm}
	\end{algorithm}

\subsubsection{A sampling-based representation approach.} The computational constraints discussed earlier can be resolved by approximating the multi-scale multimodal drug representations with a uniform truncated random walk-based sampling approach. This representation sampling technique is described by Algorithm \ref{alg:sampling_algorithm}.
The algorithm uses the bipartite graph $\mathcal{G}$ and the starting drug $d$ as input and outputs $\overline{\textbf{m}}_d$ an approximation of $\textbf{m}_d$. Before sampling starts we create $\widetilde{\textbf{m}}^0_d$ the self representation of the drug $d$ using the drug encoder $f_{\Theta_D}(\cdot)$ (line 1) and $\overline{\mathbf{m}}_d$ the vector to store the sum of the sampled multi-scale multimodal drug representations (line 2). 

We start $s$ truncated random walks on $\mathcal{G}$ starting from the drug $d$ (lines 3-4). A sampled drug representation $\widetilde{\textbf{m}}_d$ is formed by first taking the self representation $\widetilde{\textbf{m}}^0_d$. At each scale, the random walker takes a step on the bipartite graph. If the step is at an odd scale the walker samples the protein $\widetilde{p}$, a representation is created for the protein with the protein encoder $f_{\Theta_P}(\cdot)$ and $\widetilde{m}_d^l$ is concatenated to the sampled multi-scale multimodal representation (lines 7-10). When the step is at an even scale the walker samples a drug $\widetilde{d}$, a representation is distilled with the drug encoder $f_{\Theta_D}(\cdot)$ and it is concatenated to the sample vector (lines 12-14).

After the random walk is terminated, the sampled vector $\widetilde{\textbf{m}}_d$  is added to the running sum of vectors (line 17). Finally, when all of the truncated random walks have terminated $\overline{\textbf{m}}_d$ which stores the sum of the sampled multi-scale multimodal drug representations is normalized by the sample size (line 19). 

	\begin{algorithm}[h!]
{\small		\DontPrintSemicolon
		\SetAlgoLined
		\KwData{$\mathcal{B}$ -- Batch of (drug A, drug B, cell line, label) tuples.\\
		$\,\,\,\,\,\,\,\,\,\,\,\,\,\,\,\mathcal{G}$ -- Graph of interest.\\
		$\,\,\,\,\,\,\,\,\,\,\,\,\,\,\,s$ -- Sample size for multi-scale approximation.\\
		}
		\KwResult{$\mathcal{L}$ -- Average loss on batch $\mathcal{B}$.}
$\mathcal{L}\leftarrow 0 $\;
\For{$(d,d',c,y) \in \mathcal{B}$}{
$\overline{\textbf{m}}_{d}\leftarrow \textbf{Sample Representation}(\mathcal{G},d,s)$\;
$\overline{\textbf{m}}_{d'}\leftarrow \textbf{Sample Representation}(\mathcal{G},d',s)$\;
$\textbf{h}_{(d,d',c)}\leftarrow [\overline{\textbf{m}}_{d} \| \overline{\textbf{m}}_{d'} \|  \textbf{h}_{c}]$\;
$\hat{y}_{(d,d',c)} \leftarrow  f_{\Theta_H}(\textbf{h}_{(d,d',c)})$\;

$\mathcal{L} \leftarrow  \mathcal{L} + \ell(y; \hat{y}_{(d,d',c)})/|\mathcal{B}|$\;
}
}
		\caption{Mini-batch based forward pass of MOOMIN-A with approximate multi-scale multimodal representations.}\label{alg:batching}
	\end{algorithm}
\vspace{-5mm}

\subsection{Synergy scoring layer and training}
The primary purpose of MOOMIN is to predict the synergistic nature of drug pairings conditioned on the targeted cancer cell lines. We achieve this with  $f_{\theta_H}(\cdot)$ the synergy scoring layer of our model defined by Equation \eqref{eq:head_layer} which depends on the drug pair-cell representation $\textbf{h}_{(d,d',c)}$ defined by Equation \eqref{eq:drug_drug_cell}. The layer outputs $\hat{y}_{d,d',c}$ the probability that the drug pair $d, d'$ is effective at killing the cancer cell $c$. In practice, the synergy scoring layer is a feedforward neural network parametrized by $\Theta_H$. It has a single hidden layer and a final neuron with a sigmoid activation function.
\begin{align}
\textbf{h}_{(d,d',c)} &=[\textbf{m}_{d} \| \textbf{m}_{d'} \|  \textbf{h}_{c}]\label{eq:drug_drug_cell}\\
\hat{y}_{(d, d', c)} &= f_{\Theta_H}(\textbf{h}_{(d,d',c)})\label{eq:head_layer}
\end{align}
Using the synergy probability outputted by the synergy scoring layer we compute the loss for a single entry in the drug synergy database based on the binary cross-entropy in Equation \eqref{eq:logloss}.
\begin{align}
\ell(y; \hat{y}) =&-   [y \cdot  \log (\hat{y})
+(1-y) \cdot\log(1-\hat{y})]\label{eq:logloss}
\end{align}
These losses can be averaged out over the synergy database and used to calculate the mean binary cross-entropy on the synergy database $\mathcal{S}$ defined by Equation \eqref{eq:mean_loss}. 
\begin{align}
\mathcal{L}&=\sum\limits_{\mathclap{(d,d',c,y) \in \mathcal{S}}}\ell(y_{d,d',c}; \hat{y}_{d,d',c})/|\mathcal{S}|\label{eq:mean_loss}
\end{align}
The model which calculates the transition probabilities in Equation \eqref{eq:multiscale_probs} and the average loss in Equation \eqref{eq:mean_loss} exactly is referenced as MOOMIN-E. By minimizing this average loss, $\Theta_D,\Theta_P, \Theta_C$ and $\Theta_H$ the parameters of the encoders and the synergy scoring layer can be learned jointly. We also propose an efficient mini-batching-based approximate model called MOOMIN-A which uses the approximation technique described by Algorithm \ref{alg:sampling_algorithm} for creating multi-scale representations and batching procedure of Algorithm \ref{alg:batching}.

\section{Experimental Evaluation}\label{sec:experiments}
In this section, we evaluate the predictive performance of MOOMIN under various training and testing regimes. Furthermore, we provide results about the efficiency of learning, stability of the approximate inference, validate model predictions on external datasets, and analyze the hidden representations visually.

\subsection{Predictive performance}\label{subsec:predictive_performance}
The drug synergy scoring problem can be solved by a range of existing machine learning techniques. Our goal is to have an extensive predictive performance comparison with the existing set of tools. 

\subsubsection{Baselines} The benchmark methods cover a range of graph and node level representation learning algorithms. Unsupervised methods use a two-stage pipeline; (i) drug embedding generation and pairwise representation concatenation (ii) gradient boosted machine training and scoring with scikit-learn \cite{pedregosa2011scikit}. Specifically, the baseline techniques are:
\begin{itemize}
    \item \textit{Molecule statistical descriptors.} We calculate graph descriptors for each drug molecule using the default settings of the \textit{Karate Club} \cite{rozemberczki2020karate} library.
    \item \textit{Node embeddings.} We create higher order proximity-preserving node embeddings from the drug-protein interaction graph with the baseline hyperparameters of \textit{Karate Club} \cite{rozemberczki2020karate}.
    \item \textit{Tensor factorization.} We factorize the heterogeneous interaction graph with \textit{PyKeen} \cite{ali2021pykeen} and use the drug node representations as features with the default settings.
    \item \textit{Graph neural networks.} Using the molecular structure extracted from the SMILES strings we train pooled graph neural networks with PyTorch Geometric \cite{fey2019fast}.
\end{itemize}
\subsubsection{Experimental settings.} We used default hyperparameters from \cite{kipf2017semi} and \cite{predict2019klicpera} in order to ensure fair and comparable evaluation. The APPNP \textit{drug encoders} in MOOMIN uses 10 approximate Personalized PageRrank \cite{bojchevski2020scaling,predict2019klicpera} iterations with a return probability of 0.2 and has two hidden layers with 32 neurons separated by ReLU \cite{nair2010rectified} activations. Atom representations are pooled by mean, max and min pooling functions, and the graph representations are concatenated together. \textit{Protein encoders} in MOOMIN had a single hidden layer with 32 neurons and the same activation functions. Cancer \textit{cell line encoders} are 16 dimensional \textit{embeddings} initialized with the GloRot method \cite{glorot2010understanding}. The \textit{synergy prediction layer} uses 16 hidden layer neurons, ReLU activations and applies a dropout of 0.5 during training time \cite{srivastava2014dropout}. We consider truncated random walks at scales $r=\{0,1,2\}$ to assess the utility of multiple modalities and higher-order information.

The exact and approximate MOOMIN variants are optimized using the Adam optimizer \cite{kingma2015adam} and use a learning rate of $5\cdot 10^{-3}$ with a weight decay of $5\cdot 10^{-5}$. The number of samples used by MOOMIN-A is $2^7$ and the batch size is set to be $2^5$. Both MOOMIN-E and MOOMIN-A are implemented with the PyTorch and PyTorch Geometric Temporal automatic differentiation libraries \cite{paszke2019pytorch,fey2019fast}. All of the baseline models and MOOMIN variants are trained with 80\% of the drug synergy database and evaluated on the remaining 20\% of entries. We performed 5-fold cross-validation within the training set to find the optimal number of epochs based on early stopping. The mean synergy scoring performance of models calculated from 10 train/test splits is in Table \ref{tab:predictive_performance} with standard errors.

\begin{table}[h!]

\caption{The synergy scoring performance of MOOMIN, statistical fingerprinting, node embedding and deep learning approaches. We report mean performance and the standard error around the mean calculated from 10 splits. Bold numbers denote the best performing model on a metric.}\label{tab:predictive_performance}
\renewcommand{\arraystretch}{1.1}
\begin{tabular}{lccc}

\cline{2-4}

 &\textbf{ROC AUC} &\textbf{PR AUC}  & \textbf{F}$_1$ \textbf{Score}  \\
\hline
\textbf{Graph2Vec} \cite{narayanangraph2vec} & $.745\pm .002$& $.632\pm .004$& $.563\pm .002$\\
\textbf{NetLSD} \cite{tsitsulin2018netlsd,tsitsulin2020just} & $.723\pm .003$& $.604\pm .004$& $.490\pm .005$\\
\textbf{FEATHER} \cite{rozemberczki2020characteristic} & $.732\pm .003$& $.610\pm .005$& $.520\pm .004$\\
\textbf{GeoScattering} \cite{gao2019geometric} & $.726\pm .004$& $.612\pm .004$& $.499\pm .004$\\
 \midrule
\textbf{NetMF}  \cite{qiu2018network}&$.751\pm .001$& $.678\pm .006$& $.553\pm .004$\\
\textbf{GraRep}  \cite{cao2015grarep}&$.760\pm .001$& $.684\pm .006$& $.595\pm .003$\\
 \textbf{Walklets} \cite{perozzi2017don}&$.762\pm .002$& $.689\pm .001$& $.581\pm .002$\\
\textbf{LINE}  \cite{tang2015line}&$.760\pm .002$& $.686\pm .005$& $.576\pm .003$\\
\textbf{HOPE} \cite{ou2016asymmetric}&$.757\pm .003$& $.679\pm .006$& $.578\pm .005$\\
\textbf{DeepWalk}\cite{perozzi2014deepwalk}&$.761\pm .002$& $.686\pm .003$& $.595\pm .004$\\
\hline 
\textbf{GCN} \cite{kipf2016semi} &$.714 \pm .005$ & $.613 \pm .009$ & $.534 \pm .053$ \\
\textbf{GAT} \cite{graph2018velickovic} &$.700 \pm .006$ & $.593 \pm .005$ & $.563 \pm .021$ \\
\textbf{MixHop} \cite{abu2019mixhop}& $.731 \pm .002$ & $.633 \pm .004$ & $.537 \pm .066$ \\
 \hline 
\textbf{ComplEx} \cite{trouillon2017knowledge}&$.764\pm .003$ & $.691\pm .002$ & $.581\pm .003$ \\
\textbf{RESCAL} \cite{nickel2011three}& $.762\pm .002$&  $.684\pm .004$& $.577\pm .003$ \\
\textbf{TuckER} \cite{balazevic2019tucker}& $.748\pm .003$&  $.664\pm .004$& $.521\pm .003$ \\
\textbf{ComboFM} \cite{julkunen2020leveraging} & $.760 \pm .002$& $.681\pm.005 $& $.563\pm.018$\\

\textbf{DualE} \cite{cao2021dual} & $.758\pm.003$& $.683\pm.004$& $.576\pm.002$\\

\textbf{SEEK} \cite{xu2020seek} & $.755\pm.002$& $.676.\pm.003$& $.548.\pm.004$\\

\textbf{ReinceptionE} \cite{xie2020reinceptione} & $.758\pm.002$& $.685\pm.003$& $.553\pm.003$\\

\textbf{CompGCN} \cite{vashishth2019composition} & $.764\pm.003$& $.688\pm.002$& $.584\pm .0.06$\\

 \hline
 \textbf{MOOMIN-E} $r=0$&$.705\pm .003$& $.639\pm .004$& $.533\pm .002$\\
 \textbf{MOOMIN-E} $r=1$&$\textbf{.777}\mathbf{\pm} \textbf{.002}$& $.702\pm .002$& $.632\pm .003$\\
 \textbf{MOOMIN-E} $r=2$&$.770\pm .003$ &$\textbf{.708}\mathbf{\pm} \textbf{.002}$  &$\textbf{.656}\mathbf{\pm} \textbf{.003}$  \\
 \hline
  \textbf{MOOMIN-A} $r=0$&$.718\pm .003$ &  $.656\pm .005$&  $.545\pm .006$\\
 \textbf{MOOMIN-A} $r=1$& $.775\pm .004$&  $.699\pm .005$& $.588\pm .005$ \\
 \textbf{MOOMIN-A} $r=2$& $.753\pm .003$&  $.674\pm .006$& $.568\pm .008$ \\
 \bottomrule
\end{tabular}
\end{table}
\vspace{-5mm}

\subsubsection{Hyperparameters for fair evaluation}
The baselines' hyperparameters ensure a fair evaluation with respect to expressive power (number of free parameters) and the optimizer settings because:
\begin{enumerate}
    \item Tensor factorization, node embedding, and graph fingerprinting techniques have the same embedding dimensions.
    \item Drug pair classifier graph neural network models are trained with the same optimizer settings as MOOMIN.
    \item It holds at $r<3$ that MOOMIN has fewer embedding dimensions and free trainable parameters than the node embedding, graph fingerprinting, and tensor factorization techniques.
    \item When $r=0$ the drug pair classifier graph neural networks and MOOMIN have the same number of trainable parameters and the optimizer settings are the same.
\end{enumerate}

\subsubsection{Experimental findings.} Our results in Table \ref{tab:predictive_performance} demonstrate that the MOOMIN models can significantly outperform a range of strong baselines on all evaluation metrics. In terms of $F_1$ score this advantage can be as big as 7.5\%, while the gain with respect to AUC score is only 1.7\%. Our results show that incorporating protein features at the $1^{st}$ scale is a universally beneficial choice regardless of the metric used for evaluation. Higher than $1^{st}$ order information only improves the performance of MOOMIN-E when PR AUC and F$_1$ scores are utilized. Our results also validate the efficacy of MOOMIN-A in the low-order proximity regime. In fact, for $r=1$ based on the mean performance of MOOMIN-E and MOOMIN-A measured by ROC AUC and PR AUC is not significantly different.

\subsection{Tissue conditional performance}
Cancer cell lines on which the drug pairs are tested all have a source tissue from which they were originally extracted. We specifically analyze whether the predictive performance of MOOMIN-E depends on the source tissue of cancer cells.
\subsubsection{Experimental settings.} 
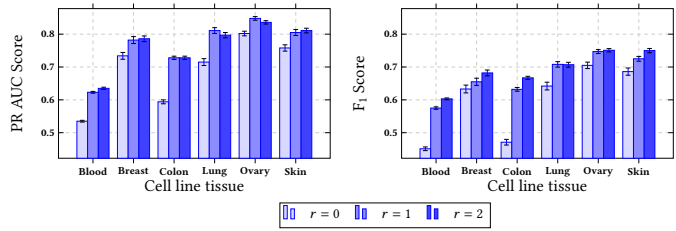
\begin{figure}[h!]

	\centering
	\begin{tikzpicture}[scale=0.45,transform shape]
	\tikzset{font={\fontsize{14pt}{12}\selectfont}}
	\begin{groupplot}[group style={group size=2 by 2,
		horizontal sep=60pt, vertical sep=60pt,ylabels at=edge left},
	width=0.54\textwidth,
	height=0.3375\textwidth,
	ymin=0.48,
	ymax=0.82,
	legend columns=3,
every tick label/.append style={font=\bf},
    y tick label style={
        /pgf/number format/.cd,
            fixed,
            fixed zerofill,
            precision=0,
        /tikz/.cd
    },
 enlarge x limits=true,
	grid=major,
	grid style={dashed, gray!40},
	scaled ticks=false,
	inner axis line style={-stealth}]

 \nextgroupplot[
   xlabel=Cell line tissue,
    ybar=0pt,
      every tick/.style={
        black,
        semithick,
      },
    bar width=9pt,
    enlargelimits=0.17,
    ylabel={PR AUC Score},
    legend style={at={(0.5,-0.15)},
      anchor=north,legend columns=-1},
yticklabels={0.5,0.6,0.7,0.8},
ytick={0.5,0.6,0.7,0.8},
    symbolic x coords={Blood,Breast,Colon,Lung,Ovary,Skin},
    xtick={Blood, Breast, Colon, Lung, Ovary, Skin},
    ]

\addplot [fill=blue!15,draw=blue,error bars/.cd,y dir=both,y explicit]  coordinates {
(Blood,0.535) +- ( 0.0027 , 0.0027 )
(Breast,0.734) +- ( 0.0103 , 0.0103 )
(Colon,0.594) +- ( 0.006 , 0.006 )
(Lung,0.715) +- ( 0.0105 , 0.0105 )
(Ovary,0.802) +-(0.007 , 0.007 )
(Skin,0.758) +- (0.0096 , 0.0096 )};
\addplot [fill=blue!45,draw=blue,error bars/.cd,y dir=both,y explicit]  coordinates {
(Blood,0.623) +- ( 0.0027 , 0.0027 )
(Breast,0.782) +- ( 0.0106 , 0.0106 )
(Colon,0.728) +- ( 0.0049 , 0.0049 )
(Lung,0.811) +- ( 0.0088 , 0.0088 )
(Ovary,0.848) +- ( 0.0055 , 0.0055 )
(Skin,0.805) +- ( 0.0091 , 0.0091 )};
\addplot [fill=blue!75,draw=blue,error bars/.cd,y dir=both,y explicit] coordinates {
(Blood,0.635) +- ( 0.0035 , 0.0035 )
(Breast,0.786) +- ( 0.0086 , 0.0086 )
(Colon,0.728) +- ( 0.005 , 0.005 )
(Lung,0.797) +- ( 0.0081 , 0.0081 )
(Ovary,0.836) +- ( 0.0051 , 0.0051 )
(Skin,0.811) +- ( 0.0071 , 0.0071 )
};

 \nextgroupplot[
   xlabel=Cell line tissue,
    ybar=0pt,
      every tick/.style={
        black,
        semithick,
      },
    bar width=9pt,
    enlargelimits=0.17,
    legend columns=4,
    legend image post style={solid},
    legend style={at={(0.5,-0.25)},nodes={scale=1.5, transform shape}, 
      anchor=north,legend columns=-1},
    ylabel={F$_{1}$ Score},
yticklabels={0.5,0.6,0.7,0.8},
ytick={0.5,0.6,0.7,0.8},
    symbolic x coords={Blood,Breast,Colon,Lung,Ovary,Skin},
    xtick={Blood,Breast,Colon,Lung,Ovary,Skin},
    	legend style = { column sep = 10pt, legend columns = 1, legend to name = grouplegend, font=\small}  ]

\addplot [fill=blue!15,draw=blue,error bars/.cd,,y dir=both,y explicit]  coordinates {
(Blood,0.451) +- ( 0.0056 , 0.0056 )
(Breast,0.633) +- ( 0.0119 , 0.0119 )
(Colon,0.471) +- ( 0.0086 , 0.0086 )
(Lung,0.642) +- ( 0.0118 , 0.0118 )
(Ovary,0.705) +- ( 0.0098 , 0.0098 )
(Skin,0.686) +- ( 0.011 , 0.011 )};\addlegendentry{$r=0$}

\addplot [fill=blue!45,draw=blue,error bars/.cd,y dir=both,y explicit]   coordinates {
(Blood,0.575) +- ( 0.0037 , 0.0037 )
(Breast,0.655) +- ( 0.0109 , 0.0109 )
(Colon,0.632) +- ( 0.0057 , 0.0057 )
(Lung,0.708) +- ( 0.0083 , 0.0083 )
(Ovary,0.747) +- ( 0.006 , 0.006 )
(Skin,0.725) +- ( 0.0073 , 0.0073 )};\addlegendentry{$r=1$}

\addplot [fill=blue!75,draw=blue,error bars/.cd,y dir=both,y explicit] coordinates {
(Blood,0.603) +- ( 0.0028 , 0.0028 )
(Breast,0.682) +- ( 0.0089 , 0.0089 )
(Colon,0.667) +- ( 0.0047 , 0.0047 )
(Lung,0.707) +- ( 0.0073 , 0.0073 )
(Ovary,0.751) +- ( 0.0051 , 0.0051 )
(Skin, 0.75) +- ( 0.0065 , 0.0065 )
};\addlegendentry{$r=2$}
	\end{groupplot}

	\node at ($(group c2r1) + (-4.5cm,-3.9cm)$) {\ref{grouplegend}}; 
	\end{tikzpicture}
	
	\caption{The synergy scoring performance of MOOMIN conditioned on cell line tissue. The bars are mean performance metrics with standard deviations around the mean calculated from 10 experimental repetitions .}\label{fig:tissue_conditioned}
\end{figure}

We utilize the hyperparameters from Subsection \ref{subsec:predictive_performance}. Using 10 experimental runs we calculate mean predictive performance metrics with standard errors around the mean on the test set conditioned on the most common cell line tissues. We plotted the average PR AUC and F$_1$ scores with the standard error bars on subplots of Figure \ref{fig:tissue_conditioned} for the proximity scales $r\in\{0,1,2\}$.
\vspace{-5mm}
\subsubsection{Experimental findings.} The bar charts on Figure \ref{fig:tissue_conditioned} show that predictive performance is cell line tissue dependent: the model performs remarkably well on lung and ovary cells and relatively poorly on blood and colon cells. Including structural information about the interacting proteins can increase the test set F$_1$ scores on blood and colon cells by as much as 10\% which is a remarkable predictive performance difference. The same performance gain pattern holds for the PR AUC based performance on the blood and colon tissue-specific cancer cell lines.

\subsection{Data efficiency}
The design of MOOMIN allows for semi-supervised learning; the characteristics of drugs that are not in the training database can be exploited to contextualize pairs of drugs used for training better. We hypothesize that because of this MOOMIN has a reasonable predictive performance in the low training data ratio regime.

\subsubsection{Experimental settings.} 
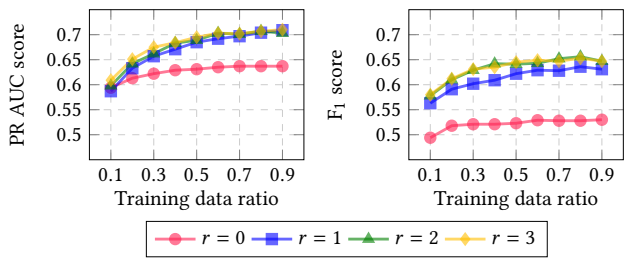
\begin{figure}[h!]
\centering
\scalebox{0.9}{
\begin{tikzpicture}
\begin{groupplot}[	grid=major,
	grid style={dashed, gray!40},group style={
                      group name=myplot,
                      group size= 2 by 1, horizontal sep=1.55cm,vertical sep=1.2cm},height=3.8cm,width=4.75cm, title style={at={(0.5,0.9)},anchor=south},every axis x label/.style={at={(axis description cs:0.5,-0.15)},anchor=north},]
\nextgroupplot[
	ylabel=PR AUC score,
	xlabel=Training data ratio,
	xtick={0.1,0.3,0.5,0.7,0.9},
	xmin=0,
	xmax=1.0,
	ymin=0.45,
	ymax=0.75,
	ytick={0.5,0.55,0.6,0.65,0.7},
 	legend columns=4,
	legend style={at={(1.20,-0.65)},anchor=south},
y label style={at={(0.05,0.5)}},
    legend entries={$r=0$, $r=1$, $r=2$,$r=3$},
]
\addplot [very thick, awesome,mark=*,opacity=0.6]coordinates {
( 0.1 , 0.594 )
( 0.2 , 0.613 )
( 0.3 , 0.622 )
( 0.4 , 0.629 )
( 0.5 , 0.631 )
( 0.6 , 0.635 )
( 0.7 , 0.637 )
( 0.8 , 0.637 )
( 0.9 , 0.637 )
};
\addplot [very thick, blue,mark=square*,opacity=0.6]coordinates {
( 0.1 , 0.587 )
( 0.2 , 0.633 )
( 0.3 , 0.657 )
( 0.4 , 0.671 )
( 0.5 , 0.685 )
( 0.6 , 0.692 )
( 0.7 , 0.697 )
( 0.8 , 0.704 )
( 0.9 , 0.709 )
};
\addplot [very thick, ao(english), ,mark=triangle*,opacity=0.6]coordinates {
( 0.1 , 0.598 )
( 0.2 , 0.642 )
( 0.3 , 0.662 )
( 0.4 , 0.683 )
( 0.5 , 0.689 )
( 0.6 , 0.703 )
( 0.7 , 0.701 )
( 0.8 , 0.707 )
( 0.9 , 0.703 )
};
\addplot [very thick, amber,mark=diamond*,opacity=0.6]coordinates {
( 0.1 , 0.609 )
( 0.2 , 0.651 )
( 0.3 , 0.675 )
( 0.4 , 0.684 )
( 0.5 , 0.695 )
( 0.6 , 0.703 )
( 0.7 , 0.702 )
( 0.8 , 0.707 )
( 0.9 , 0.71 )
};

\nextgroupplot[
	xtick={0.1,0.3,0.5,0.7,0.9},
    y label style={at={(0.05,0.5)}},
	ylabel=F$_1$ score,
	xlabel=Training data ratio,
	xmin=0,
	xmax=1.0,
	ymin=0.45,
	ymax=0.75,
	ytick={0.5,0.55,0.6,0.65,0.7},
	]
\addplot [very thick, awesome,mark=*,opacity=0.6]coordinates {
( 0.1 , 0.494 )
( 0.2 , 0.518 )
( 0.3 , 0.521 )
( 0.4 , 0.521 )
( 0.5 , 0.523 )
( 0.6 , 0.529 )
( 0.7 , 0.528 )
( 0.8 , 0.528 )
( 0.9 , 0.53 )
};
\addplot [very thick, blue,mark=square*,opacity=0.6]coordinates {
( 0.1 , 0.563 )
( 0.2 , 0.591 )
( 0.3 , 0.602 )
( 0.4 , 0.609 )
( 0.5 , 0.622 )
( 0.6 , 0.629 )
( 0.7 , 0.628 )
( 0.8 , 0.636 )
( 0.9 , 0.631 )
};
\addplot [very thick, ao(english), ,mark=triangle*,opacity=0.6]coordinates {
( 0.1 , 0.578 )
( 0.2 , 0.61 )
( 0.3 , 0.629 )
( 0.4 , 0.642 )
( 0.5 , 0.641 )
( 0.6 , 0.643 )
( 0.7 , 0.652 )
( 0.8 , 0.656 )
( 0.9 , 0.647 )
};
\addplot [very thick, amber,mark=diamond*,opacity=0.6]coordinates {
( 0.1 , 0.58 )
( 0.2 , 0.612 )
( 0.3 , 0.631 )
( 0.4 , 0.635 )
( 0.5 , 0.645 )
( 0.6 , 0.649 )
( 0.7 , 0.647 )
( 0.8 , 0.653 )
( 0.9 , 0.647 )
};

\end{groupplot}
\end{tikzpicture}}
\caption{The mean synergy scoring performance of MOOMIN-E conditioned on the ratio of training data calculated from 10 experimental runs. }\label{fig:data_ratio}
\end{figure}

We train MOOMIN-E with the hyperparameter settings detailed in Subsection \ref{subsec:predictive_performance}. For each training data ratio, we do 10 splits and calculate the average predictive performance on the test set. These mean performance scores are plotted on subplots of Figure \ref{fig:data_ratio} as a function of training data ratio for the proximity scales $r\in\{0,1,2,3\}$.
\vspace{-5mm}

\subsubsection{Experimental findings.} The line charts in Figure \ref{fig:data_ratio} demonstrate that MOOMIN is efficient at utilizing the training data as the average test set performance increases steadily. In terms of PR AUC and F$_1$ scores, the higher order models can outperform baselines by using half of the training data when Table \ref{tab:predictive_performance} and Figure \ref{fig:data_ratio} are cross-referenced. It is also evident that the purely molecular feature based $0^{th}$ scale model has moderate gains with additional training data compared to true multimodal models.
\begin{table}[h!]
\centering
\caption{The molecule and window size conditional test set performance of MOOMIN-E. We report average predictive performances calculated from 10 experimental runs with standard errors around the mean.}
\setlength{\tabcolsep}{4pt}
{\small
\begin{tabular}{ccccc}


 \textbf{Molecule Pair}&\textbf{Window}&\textbf{ROC AUC} &\textbf{PR AUC}  & \textbf{F}$_1$ \textbf{Score} \\
\hline
 &0&$.724\pm.007$ & $.767\pm .007$ & $.676 \pm .022$  \\
Large - Large &1&$\mathbf{.785\pm .007}$ & $\mathbf{.808\pm .008}$ & ${.733\pm .010}$  \\
 &2&${.763\pm .013}$ & ${.803\pm .008}$ & $\mathbf{.745\pm .006}$  \\
\hline
 &0&${.713\pm.005}$ & ${.656\pm .008}$ & ${.543 \pm .022}$  \\
Large - Small &1&$\mathbf{.778\pm .005}$ & ${.713\pm .008}$ & ${.656\pm .012}$  \\
 &2&${.768\pm .006}$ & $\mathbf{.717\pm .012}$ & $\mathbf{.678\pm .007}$  \\
\hline
 &0&${.614\pm .008}$ & ${.447\pm .011}$ & ${.312\pm .016}$  \\

Small - Small &1&${.715\pm .006}$ & ${.554\pm .010}$ & ${.477\pm .016}$  \\

 &2&$\mathbf{.716\pm .007}$ & $\mathbf{.563\pm .010}$ & $\mathbf{.516\pm .013}$  \\
 \bottomrule
\end{tabular}}
\end{table}
\vspace{-5mm}
\subsection{Molecule size conditional performance}
The drugs in DrugCombDB are heterogeneous concerning the number of atoms they contain and the pairings sometimes have drugs that are considerably different concerning size.

\subsubsection{Experimental settings.} Using the MOOMIN-E test set synergy scores obtained in  Subsection \ref{subsec:predictive_performance} we calculate molecule size conditional performance metrics for various window sizes. We used the following definitions to categorize the drug pairs based on the molecule sizes for calculating the conditional performance metrics: (i) \textit{Large - Large}: If both molecules have 50 or more non-hydrogen atoms; (ii) \textit{Large - Small}: If only one of the molecules has 50 or more non-hydrogen atoms; (iii) \textit{Small - Small}: If both molecules have less than 50 non-hydrogen atoms.

\subsubsection{Experimental findings.} Our results show that pairs with two small molecules gain the most in relative terms by adding the protein feature-based context. These pairs have gains even when the maximal window size is increased to 2. This heterogeneity of performances concerning the molecule size and maximal scale being considered implies that molecule sizes specific models should be trained. For example, small molecule pairs should be scored with a MOOMIN model with a large maximal scale.

\subsection{Inference stability}
The MOOMIN-A variant uses sampled truncated random walks to approximate the transition probabilities between the vertices in the interaction graph. Inference also uses a sampling-based approach -- the variance of the test set performance is affected by sample size.
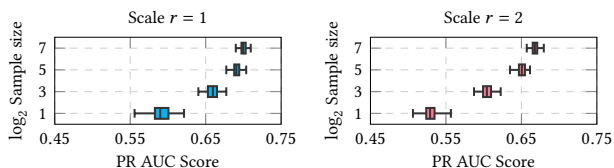
\begin{figure}[h!]
	\centering
	\begin{tikzpicture}[scale=0.95,transform shape]
	\tikzset{font={\fontsize{7pt}{8}\selectfont}}
\begin{groupplot}[
	grid=major,
	grid style={dashed, gray!40},
group style={group name=myplot,
                      group size= 2 by 1, horizontal sep=1.25cm,vertical sep=1.15cm},height=2.8cm,width=4.75cm, title style={at={(0.5,0.9)},anchor=south},
                     xlabel style={at={(0.5,-0.15)},anchor=south},
                      ylabel style={at={(0.3,0.5)},anchor=south},]

\nextgroupplot[
    ytick={1,2,3,4},
    ylabel=$\log_2$ Sample size,
    xlabel=PR AUC Score,  yticklabels={1,3,5,7},
    ymin=0.5,
    ymax=4.5,
    title={Scale $r=1$},
    xmin=0.45,
    xmax=0.75,
    xtick={0.45, 0.55,0.65,0.75}
 ]

    \addplot+[thick,solid,
    black!80,
    fill=cyan!90,mark={diamond*},
    boxplot prepared={
        box extend=0.5, 
lower whisker=0.5558,
lower quartile=0.5827,
median=0.5898,
upper quartile=0.6007,
upper whisker=0.6213
    },
    ] coordinates {};
    \addplot+[thick,solid,
    black!80,
    fill=cyan!90,mark={diamond*},
    boxplot prepared={
        box extend=0.5, 
lower whisker=0.6403,
lower quartile=0.6529,
median=0.6596,
upper quartile=0.665,
upper whisker=0.6773
    },
    ] coordinates {};
    \addplot+[thick,solid,
    black!80,
    fill=cyan!90,mark={diamond*},
    boxplot prepared={
        box extend=0.5, 
lower whisker=0.6773,
lower quartile=0.6878,
median=0.6906,
upper quartile=0.6945,
upper whisker=0.7037
    },
    ] coordinates {};

    \addplot+[thick,solid,
    black!80,
    fill=cyan!90,mark={diamond*},
    boxplot prepared={
        box extend=0.5, 
lower whisker=0.6899,
lower quartile=0.6974,
median=0.6996,
upper quartile=0.703,
upper whisker=0.7097
    },
    ] coordinates {};

\nextgroupplot[
    ytick={1,2,3,4},
    ylabel=$\log_2$ Sample size,
    xlabel=PR AUC Score,
    yticklabels={1,3,5,7},
    title={Scale $r=2$},
    ymin=0.5,
    ymax=4.5,
    xmin=0.45,
    xmax=0.75,
    xtick={0.45,0.55,0.65,.75}
 ]
    \addplot+[thick,solid,
    black!80,
    fill=awesome!60,mark={diamond*},
    boxplot prepared={
        box extend=0.5, 
lower whisker=0.5063,
lower quartile=0.5239,
median=0.5290,
upper quartile=0.5349,
upper whisker=0.5566
    },
    ] coordinates {};
    \addplot+[thick,solid,
    black!80,
    fill=awesome!60,mark={diamond*},
    boxplot prepared={
        box extend=0.5, 
lower whisker=0.5872,
lower quartile=0.5992,
median=0.6042,
upper quartile=0.6098,
upper whisker=0.6226
    },
    ] coordinates {};
    
    \addplot+[thick,solid,
    black!80,
    fill=awesome!60,mark={diamond*},
    boxplot prepared={
        box extend=0.5, 
lower whisker=0.6349,
lower quartile=0.6467,
median=0.6506,
upper quartile=0.6547,
upper whisker=0.6614
    },
    ] coordinates {};
    
    \addplot+[thick,solid,
    black!80,
    fill=awesome!60,mark={diamond*},
    boxplot prepared={
        box extend=0.5, 
lower whisker=0.6571,
lower quartile=0.6650,
median=0.6679,
upper quartile=0.6710,
upper whisker=0.6797
    },
    ] coordinates {};
    
\nextgroupplot[
    ytick={1,2,3,4},
    ylabel=$\log_2$ Sample size,
    xlabel=F$_1$ Score,  yticklabels={1,3,5,7},
    title={Scale $r=1$},
    ymin=0.5,
    ymax=4.5,
    xmin=0.4,
    xmax=0.6,
    xtick={0.4,0.45,0.5,0.55,0.6}
 ]

\end{groupplot}
\end{tikzpicture}
\caption{The stability of predictive performance under sampling based approximate inference with MOOMIN-A. Each boxplot represents the distribution of predictive performance scores calculated from 100 inference runs.}\label{fig:sampling_inference}
\end{figure}
\vspace{-5mm}
\subsubsection{Experimental settings.} We trained MOOMIN-A with the hyperparameter settings detailed in Subsection \ref{subsec:predictive_performance} at proximity scales $r\in\{1,2\}$. Using the trained model we make a 100 sampling-based predictions on the test set with $s\in \{2^1,2^3,2^5,2^7\}$. For each prediction, we calculate PR AUC scores and visualized the distribution of predictive performance metrics on subplots of Figure \ref{fig:sampling_inference} conditioned on the sample size.

\subsubsection{Experimental findings.} Looking at Figure \ref{fig:sampling_inference} we can conclude that increasing the number of truncated random walk samples increases the average predictive performance and reduces the variance of performance. The marginal average performance gains are decreasing with the sample size, but the order of decline depends on the resolution considered. Our results also support that the second-scale model could benefit from increasing the sample size at training and inference time.

\subsection{Out of sample model validation}\label{section:out_of}
Data sparsity is one of the main motivations for designing MOOMIN, so it can predict new synergy scores. Using out-of-sample combinations and the NCI Synergy Score Almanach \cite{holbeck2017national} AstraZeneca oncologists validated a handful of pairings with the cancer drug Vemurafenib \cite{laherty2011vemurafenib}.

\subsubsection{Experimental settings.} We use a MOOMIN-E model with $r=1$ and the hyperparameter settings from Subsection \ref{subsec:predictive_performance} and train the model on the whole drug synergy database. We scored all of the triples which involved the drug \textit{Vemurafenib} \citep{laherty2011vemurafenib} -- the predictions were sorted by the synergy scores and the 20 highest and lowest scoring triples were selected. Out of these combinations and cell targets, we listed seven in Table \ref{tab:samples} which are present in the NCI Almanach \citep{holbeck2017national} but are not included in the training database. 

\begin{table}[h!]
\centering
\footnotesize
\renewcommand{\arraystretch}{1.2}
\caption{Out of sample drug combination predictions on (Vemurafenib, Drug, Cell line) triples that can be externally validated in the NCI Almanach \cite{holbeck2017national}.}\label{tab:samples}
\begin{tabular}{c c c c c ccc}
\toprule

\textbf{Drug A}&\textbf{Drug B}&  \textbf{Cell line}&\textbf{Tissue}& \textbf{Score}& \textbf{Label}\\ \hline
\multirow{7}{*}{Vemurafenib}&Cabazitaxel         &MDA-MB-435&breast&.831&synergy\\
&Pazopanib         &NCI-H522&lung&.803&synergy\\
&Crizotinib          &UO-31&kidney&.807&synergy\\
&Raloxifene   &SNB-75&brain&.860&synergy\\
&Axitnib      &MDA-MB-231&breast&.043&antagonism\\
&Carboplatin   &LOX IMVI&skin&.158&antagonism\\
&Chlorambucil        &SR&blood&.022&antagonism\\
\bottomrule
\end{tabular}

\end{table}

\vspace{-5mm}
\subsubsection{Experimental findings.} Out of sample predictions in Table \ref{tab:samples} which were manually validated come from a diverse set of drugs and cell lines that are derived from multiple source tissues (e.g kidney and brain). This shows that MOOMIN can identify new synergistic drug combinations without doing the actual experiments.

\subsection{Runtime}
\begin{figure}[h!]
\centering
\scalebox{0.8}{
\begin{tikzpicture}
\begin{groupplot}[	grid=major,
	grid style={dashed, gray!40},group style={
                      group name=myplot,
                      group size= 2 by 1, horizontal sep=0.75cm,vertical sep=0.75cm},height=3.8cm,width=4.75cm, title style={at={(0.5,0.9)},anchor=south},every axis x label/.style={at={(axis description cs:0.5,-0.15)},anchor=north},]
\nextgroupplot[
    title={Scale $r=1$},
	ylabel=Relative runtime \%,
	xlabel=$\log_2$ Batch size,
	xtick={1,3,5,7},
	xmin=0.0,
	xmax=8.0,
	ymin=-15.0,
	ymax=115.0,
	ytick={0,25,50,75,100},
 	legend columns=1,
	legend style={at={(2.70,-0.0)},anchor=south},
y label style={at={(0.05,0.5)}},
    legend entries={$s=2^1$, $s=2^3$, $s=2^5$,$s=2^7$},
]
\addplot [very thick, awesome,mark=*,opacity=0.6]coordinates {
(1 , 1.396 )
(2 , 1.509 )
(3 , 1.849 )
(4 , 2.465 )
(5 , 3.672 )
(6 , 5.576 )
(7 , 10.919)};
\addplot [very thick, blue,mark=square*,opacity=0.6]coordinates {
( 1 , 1.553 )
( 2 , 1.887 )
( 3 , 2.552 )
( 4 , 3.569 )
( 5 , 5.489 )
( 6 , 11.158 )
( 7 , 22.373 )
};
\addplot [very thick, ao(english), ,mark=triangle*,opacity=0.6]coordinates {
( 1 , 2.03 )
( 2 , 2.503 )
( 3 , 3.479 )
( 4 , 5.356 )
( 5 , 10.667 )
( 6 , 22.04 )
( 7 , 46.738 )
};
\addplot [very thick, amber,mark=diamond*,opacity=0.6]coordinates {
( 1 , 2.584 )
( 2 , 3.61 )
( 3 , 5.281 )
( 4 , 10.391 )
( 5 , 21.603 )
( 6 , 46.422 )
( 7 , 106.722 )
};

\nextgroupplot[
    y label style={at={(0.05,0.5)}},
    title={Scale $r=2$},
	ylabel={},
	xlabel=$\log_2$ Batch size,
	xtick={1,3,5,7},
	xmin=0.0,
	xmax=8.0,
	ymin=-15.0,
	ymax=115.0,
	ytick={0,25,50,75,100},
	]
\addplot [very thick, awesome,mark=*,opacity=0.6]coordinates {
( 1 , 1.203 )
( 2 , 1.333 )
( 3 , 1.818 )
( 4 , 2.322 )
( 5 , 3.22 )
( 6 , 6.018 )
( 7 , 10.423 )
};
\addplot [very thick, blue,mark=square*,opacity=0.6]coordinates {
( 1 , 1.359 )
( 2 , 1.765 )
( 3 , 2.244 )
( 4 , 3.195 )
( 5 , 5.447 )
( 6 , 10.435 )
( 7 , 21.577 )
};
\addplot [very thick, ao(english), ,mark=triangle*,opacity=0.6]coordinates {
( 1 , 1.793 )
( 2 , 2.295 )
( 3 , 3.224 )
( 4 , 5.222 )
( 5 , 11.185 )
( 6 , 21.659 )
( 7 , 46.993 )
};
\addplot [very thick, amber,mark=diamond*,opacity=0.6]coordinates {
( 1 , 2.3 )
( 2 , 3.236 )
( 3 , 5.281 )
( 4 , 10.577 )
( 5 , 21.654 )
( 6 , 46.833 )
( 7 , 109.679 )
};

\end{groupplot}
\end{tikzpicture}}
\caption{The relative runtime of a forward and backpropagation pass with MOOMIN-A compared to MOOMIN-E.}\label{fig:relative_runtime}
\end{figure}
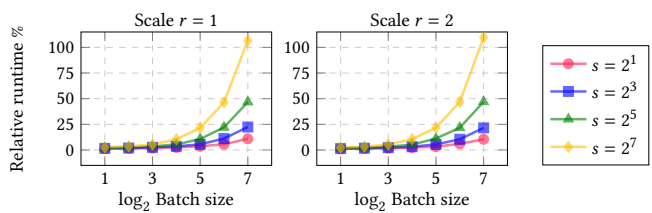

The whole dataset used for training the MOOMIN variants is fairly small in terms of drug combinations. Training and scoring the exact or approximate model end-to-end on a Tesla V-100 GPU using all of the drug pairings \textit{takes less than 2 seconds}.  We investigate the relative runtime required by MOOMIN-A to make a weight update step using a single data batch. The average relative runtime for scales $r\in\left\{1,2\right\}$ calculated from 100 repetitions is visualized on the subplots of Figure \ref{fig:relative_runtime}. When the batch size and sampling count values are small MOOMIN-A has a material advantage over MOOMIN-E, the runtime values converge for larger batch and sample sizes. 
\vspace{-5mm}

\subsection{Visualizing representations}
The architecture of MOOMIN consists of separate encoders and a downstream synergy scoring head layer which are trained jointly for solving the scoring task optimally. 
\subsubsection{Experimental settings.} Using a fully trained MOOMIN-E model we extract the drug pair-cell and scoring head layer internal representations on the test set. We learn two-dimensional embeddings with the scikit-learn implementation \cite{pedregosa2011scikit} of t-SNE \cite{van2008visualizing,wattenberg2016use} and visualized the location of drug pairs on Figure \ref{fig:scatter} and colored the points by class membership -- synergistic and antagonistic pairings. 
\subsubsection{Experimental findings.} The results in Figure \ref{fig:scatter} demonstrate that the separation of synergistic and antagonistic drug pairs starts before representations are passed through the scoring layer. Embeddings based on the synergy scoring layer show a clean class conditional separation of pairs. These findings suggest that multi-scale representations created with MOOMIN could allow transfer learning to other tasks e.g. polypharmacy side effect prediction.
\input{figures/scatter_representations}

\section{Conclusions and Future Directions}\label{sec:conclusions}

In this paper, we proposed the molecular omics network (MOOMIN), a multi-modal heterogeneous graph neural network custom-tailored to predict the synergistic and antagonistic nature of drug combinations in oncology. The model consists of protein feature and molecular structure encoders which contextualize drugs at multiple scales using a bipartite drug-protein interactions graph. Using the multi-scale representations and learned cancer cell embeddings a head layer outputs synergy scores for drug pairs conditioned on the cancer cell lines. The encoders, cancer cell embeddings, and the scoring head layer were trained jointly by minimizing a binary cross-entropy-based custom loss designed for the synergy scoring task. We introduced an approximation algorithm that makes memory-efficient model training/inference possible.

Our empirical evaluation of MOOMIN which used a publicly available drug combination synergy database had shown that the predictive performance of our framework outperforms state-of-the-art proximity-based node embedding, tensor factorization, statistical fingerprinting, and deep learning methods on the synergy scoring task. We also supported evidence that learning multi-modal representations that can distill information from molecular and protein structures improves predictive performance. Our results demonstrated that this gain holds even when the performance is conditioned on the tissue of the cell lines or the amount of training data is reduced by magnitudes. Further experiments characterized approximate inference and internal representations.

We are excited about the potential future developments of our work. The enrichment of the multi-modal data is one possibility. One avenue is considering other heterogeneous biological graphs which include other node types such as biological pathways that could generate better quality multi-scale representations. Another direction is the incorporation of advanced chemical and biological features such as molecular fingerprints \cite{duvenaud2015convolutional} or information about the secondary and tertiary structure of proteins \cite{liu2018learning}.

\bibliographystyle{ACM-Reference-Format}
\bibliography{main}

\end{document}